\documentclass[accepted]{uai2026} 
                        
\usepackage[american]{babel}

\usepackage{natbib} 
\usepackage{mathtools} 
\usepackage{booktabs} 
\usepackage{tikz} 

\usepackage{times}
\usepackage{soul}
\usepackage{url}
\usepackage{hyperref}
\usepackage{inputenc}
\usepackage{caption}
\usepackage{graphicx}
\usepackage{amsmath}
\usepackage{amsthm}
\usepackage{booktabs}
\usepackage{algorithm}
\usepackage[noend]{algorithmic}
\usepackage[switch]{lineno}
\usepackage{amssymb}
\usepackage{booktabs}
\usepackage{todonotes}
\usepackage{multirow}
\usepackage{multicol}

\newtheorem{example}{Example}
\newtheorem{theorem}{Theorem}
\newtheorem{lemma}{Lemma}
\newtheorem{definition}{Definition}
\newtheorem{proposition}{Proposition}

\newcommand{\concepts}{\ensuremath{N_C}}

\newcommand{\roles}{\ensuremath{N_R}}
\newcommand{\interp}{\ensuremath{\mathcal{I}}}
\newcommand{\intdom}{\ensuremath{\Delta^\mathcal{I}}}
\newcommand{\intmap}{\ensuremath{\cdot^\mathcal{I}}}

\newcommand{\tbox}{\ensuremath{\mathcal{T}}}

\renewcommand{\Box}{\operatorname{Box}}
\newcommand{\contains}{\operatorname{Disjoint}}

\title{Approximating Probabilistic Inference in Statistical $\mathcal{EL}$ with Knowledge Graph Embeddings}

\author{%
Yuqicheng~Zhu\textsuperscript{1,2},
Nico~Potyka\textsuperscript{3},
Bo~Xiong\textsuperscript{4},
Trung-Kien~Tran\textsuperscript{1},
Mojtaba~Nayyeri\textsuperscript{2},
Evgeny~Kharlamov\textsuperscript{1,5},
Steffen~Staab\textsuperscript{2,6}
\\
\textsuperscript{1}Bosch Center for AI,
\textsuperscript{2}University of Stuttgart, 
\textsuperscript{3}Cardiff University,
\textsuperscript{4}Stanford University,\\
\textsuperscript{5}University of Oslo,
\textsuperscript{6}University of Southampton
}
  
\begin{document}
\maketitle

\begin{abstract}
  In domains where statistical data is collected across hierarchically organized categories, drawing valid conclusions requires reasoning jointly about proportions and the structure of the domain. Statistical $\mathcal{EL}$ ($\mathcal{SEL}$) formalizes this kind of reasoning, but exact inference is \textsc{ExpTime}-hard and no implementation exists. We show how knowledge graph embeddings can approximate $\mathcal{SEL}$ inference efficiently. We prove analytical runtime and soundness guarantees, and empirically evaluate the runtime and approximation quality of our approach.
\end{abstract}

\section{Introduction}
Statistical information pervades decision-making in medicine, public policy, and science, yet drawing valid conclusions from it is surprisingly error-prone \citep{thiese2015misuse, brown2018issues}. A striking example is Simpson's paradox \citep{Simpson1951}: aggregate statistics can suggest one conclusion while every relevant subgroup supports the opposite. In the 1973 UC Berkeley admissions data \citep{bickel1975sex}, women appeared to be admitted at a lower overall rate than men, yet within most individual departments the pattern disappeared or even reversed. The paradox arises because the aggregate rate conflates department-level selectivity with the distribution of male and female applicants across departments. Resolving it requires reasoning jointly about statistical proportions and the structure of the domain: what subgroups exist (departments, gender), how they relate to broader categories (all applicants), and how individuals are distributed across them.

Statistical $\mathcal{EL}$ ($\mathcal{SEL}$) \citep{penaloza2017towards} provides a formal framework for exactly this combination of hierarchical structure and statistical proportions. It represents statistical statements as probabilistic conditionals $(D \mid C)[l,u]$, expressing that the proportion of $C$-instances that also belong to $D$ lies in $[l,u]$. In the Berkeley setting, one can encode the department-specific admission rate of female applicants, e.g.\ $(\textit{Admitted} \mid \textit{DeptA\_Applicant} \sqcap \textit{Woman})[l_1,u_1]$, alongside the share of female applicants applying to each department, e.g.\ $(\textit{DeptA\_Applicant} \mid \textit{Woman})[l_2,u_2]$. Given such constraints, $\mathcal{SEL}$ can automatically derive tight bounds on the overall rate $(\textit{Admitted} \mid \textit{Woman})$, which can reveal when aggregate and subgroup-level comparisons are consistent or potentially divergent.
We demonstrate this concretely in Appendix~\ref{app:simpson}.
Further use cases arise in domains such as medicine and manufacturing, where statistics about different kinds of patients or products are reconciled across multiple levels of a hierarchy \citep{zheng2017ontology, DBLP:conf/semweb/ZhengZZKSK22}.

Despite its practical relevance, exact reasoning in $\mathcal{SEL}$ is \textsc{ExpTime}-hard \citep{Bednarczyk21}.
The only theoretically exact approach constructs an exponentially large linear program via a type elimination procedure \citep{LutzS10}, but to the best of our knowledge it has never been implemented.
$\mathcal{SEL}$ thus provides the right formalism for statistical reasoning over structured domains but has so far remained without tool support.

In this paper, we bridge this gap using knowledge graph embeddings. We embed $\mathcal{SEL}$ concepts as axis-aligned boxes in a vector space, with volumes encoding statistical proportions, and estimate unknown conditional probabilities via volume ratios, replacing intractable logical inference with efficient geometric computation. We prove that inference runs in linear time and is sound when the embedding loss is zero. In practice, the loss will typically be non-zero, but we hypothesize that the inference error is proportional to the embedding error. To evaluate this conjecture empirically, we construct three $\mathcal{SEL}$ ontologies from YAGO3 \citep{mahdisoltani2014yago3} and evaluate the runtime, approximation error of inferred probabilities, and how it relates to the embedding error. Our main contributions are:
\begin{itemize}
    \item We explain how knowledge graph embeddings can be used to approximate probabilistic inference efficiently.
    \item We apply the idea to the $\mathcal{EL}$ embedding BoxEL \citep{XiongPTNS22}, generalize embedding soundness results to $\mathcal{SEL}$ and prove novel
     runtime and soundness guarantees for approximate inference with $\mathcal{SEL}$ embeddings.
    \item We empirically evaluate the runtime,
    accuracy and approximation quality of our approach.
\end{itemize}

\section{Related Work}
\label{sec:related_work}

Probabilistic first-order logics can be classified as type 1 (statistical), type 2 (subjective) or type 3 (combined) logics \citep{halpern1990analysis}. Type 1 probabilities represent proportions in the domain, while type 2 probabilities represent a degree of belief that a statement is true. 
To the best of our knowledge, most probabilistic DLs are type 2 logics, for example,  \cite{LukasiewiczS08,Gutierrez-Basulto17,PenalozaP16}. The only other type 1 probabilistic DL that we are aware of has been sketched in the appendix of \cite{LutzS10}. Consistency-checking is ExpTime-hard for this logic as well, and we are unaware of any implementations.

Theoretically, probabilistic reasoning
in $\mathcal{SEL}$ can be performed exactly by constructing
a linear optimization problem, where the numerical variables
and constraints are derived from a type elimination procedure
and the conditionals in the knowledge base \cite{LutzS10}. 
However, this approach
is difficult to implement and not practical because the 
number of types rapidly explodes. A more pragmatic approach
for classical inference would be creating a proof system
for $\mathcal{SEL}$ similar to those for
propositional probabilistic logics \citep{FrischH94,HunterP23}.
However, even in propositional logic, completeness
of such proof systems could only be shown for very limited fragments \citep{FrischH94}.

Knowledge graph (KG) embeddings \citep{BordesUGWY13} map entities and relations into a vector space. Prominent examples include \emph{translational}  \citep{BordesUGWY13} and \emph{bilinear} embeddings \citep{yang2014embedding,DBLP:conf/icml/LiuWY17}. Many KG embeddings encode only instance-level knowledge
but 
cannot represent logical axioms. \cite{kulmanov2019embeddings} proposed embedding $\mathcal{EL}$ concepts as $n$-balls.
BoxEL \citep{XiongPTNS22} and ELBE \citep{DBLP:journals/corr/abs-2202-14018} use axis-parallel boxes instead, which is 
beneficial as they are closed under intersection (conjunction). 
Box2EL \citep{box2el} further
embeds both concepts and roles as boxes. Other types of DLs like $\mathcal{ALC}$ were also considered \citep{ozccep2020cone,garg2019quantum}. 

\section{Background on Statistical $\mathcal{EL}$}

The DL $\mathcal{EL}$ \citep{Baader03} describes concepts and their relationships using a set $\concepts$ of \emph{concept names} and a set $\roles$ of \emph{role/relation names}. Every concept name $A \in \concepts$ and the symbol $\top$ (called \emph{top}) are \emph{(atomic) $\mathcal{EL}$ concepts}. If $C_1, C_2$ are $\mathcal{EL}$ concepts and $r \in \roles$, then so are $C_1 \sqcap C_2$ and $\exists r. C$. An $\mathcal{EL}$ interpretation $\interp = (\intdom, \intmap)$ consists of a non-empty set $\intdom$ called the domain of $\interp$ and a mapping $\intmap$ that maps every concept name to a subset of $\intdom$, and every relation to a relation over $\intdom \times \intdom$. $\intmap$ is extended to arbitrary concepts as follows. We let $\top^\interp=\intdom$, $(C_1 \sqcap C_2)^\interp = C_1^\interp \cap C_2^\interp$ and $(\exists r.C)^\interp = \{a \in \intdom \mid \exists b \in C^{\interp}: (a,b) \in r^{\interp}\}$.
An $\mathcal{EL}$ \emph{TBox} contains \emph{general concept inclusions (GCIs)} of the form $C \sqsubseteq D$, where $C, D$ are concepts. An interpretation  $\interp$ satisfies $C \sqsubseteq D$ iff $C^\interp \subseteq D^\interp$. We write $C \equiv D$ as a shorthand for the two GCIs $C \sqsubseteq D$ and $D \sqsubseteq C$.

Statistical $\mathcal{EL}$, $\mathcal{SEL}$ for short, is a probabilistic extension of $\mathcal{EL}$ that allows reasoning about statistical statements \citep{penaloza2017towards}. The basic syntactic elements are \emph{(probabilistic) conditionals} $(D \mid C)[l,u]$, where $C, D$ are $\mathcal{EL}$ concept descriptions and $l,u \in[0,1] \cap \mathbb{Q}$ are (rational) probabilities such that $l \leq u$. If $l=u$, we just write $(D \mid C)[p]$. 
The intuitive reading of $(D \mid C)[l,u]$ is that the conditional probability of $D$ given $C$ is between $l$ and $u$. The probabilities are interpreted statistically. That is, $(D \mid C)[l,u]$ means that the proportion of elements in $C$ that also belong to $D$ is between $l$ and $u$. 
The formal semantics of $\mathcal{SEL}$ is defined with respect to $\mathcal{EL}$ interpretations $\interp$ with finite domain $\intdom$. $\interp$ satisfies $(D \mid C)[l,u]$,
denoted as $\interp \models (D \mid C)[l,u]$, iff either $C^\interp = \emptyset$ or $\frac{|(D \sqcap C)^\interp|}{|C^\interp|} \in [l,u]$. This is equivalent to saying that $\interp$ satisfies $(D \mid C)[l,u]$ iff
\begin{equation}
\label{eq_sel_sat}
    l \cdot |C^\interp| \leq |(D \sqcap C)^\interp|
\leq u \cdot |C^\interp|.
\end{equation}
$\mathcal{SEL}$ generalizes $\mathcal{EL}$ in the following sense \citep[Proposition 4]{penaloza2017towards}.
\begin{lemma}
\label{lemma_subs_and_eq}
For all $\mathcal{EL}$ interpretations $\interp$, we have $\interp \models C \sqsubseteq D$ iff $\interp \models (D \mid C)[1]$.
\end{lemma}
The consistency problem for $\mathcal{SEL}$ turned out to be EXP-complete \citep{BaaderE17,Bednarczyk21}. Here, we are 
interested in the following \emph{inference problem} for $\mathcal{SEL}$: Given an $\mathcal{SEL}$ TBox $\mathcal{T}$ and a query $(D \mid C)$, where $C, D$ are
$\mathcal{SEL}$ concepts, find the largest lower bound $l$ and the smallest upper bound $u$ such that every interpretation
$\interp$ that satisfies $\mathcal{T}$ also satisfies
$(D \mid C)[l,u]$. 

\section{Illustrative Example}\label{sec:example}
To foster intuition, we illustrate how embeddings can approximate statistical reasoning in $\mathcal{SEL}$ using a simplified university admissions scenario inspired by the Berkeley example from the introduction.

\begin{figure}[h!]
\centering
\includegraphics[width=.45\textwidth]{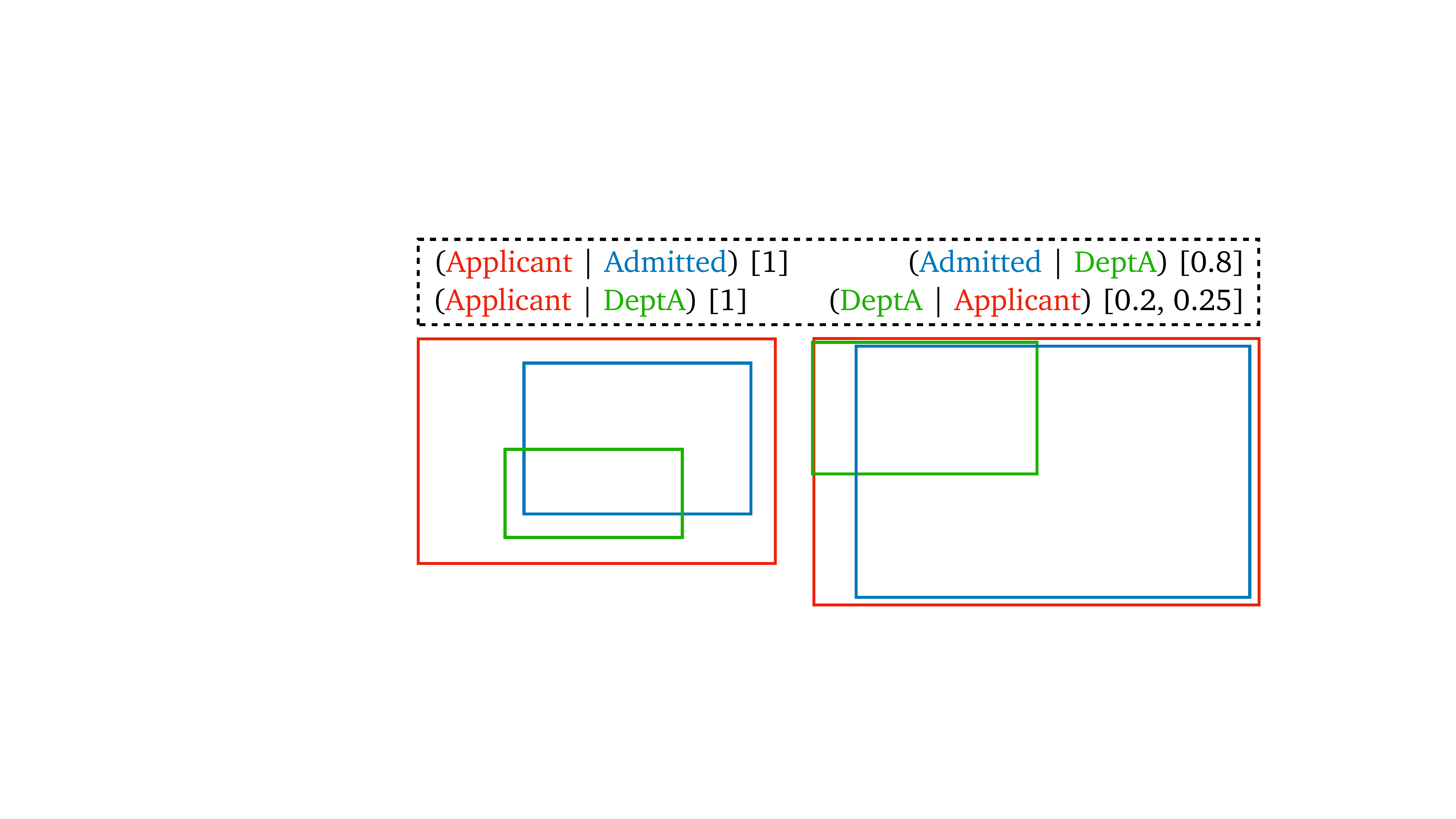}
\caption{Two possible 2D box embeddings of the concepts Applicant (red), Admitted (blue) and DeptA (green) that maintain proportions stated in the knowledge base.}\label{fig:proportions}
\end{figure}

\begin{example}
Figure~\ref{fig:proportions} shows, at the top, some $\mathcal{SEL}$ axioms.
Every admitted applicant and every Department~A applicant (DeptA) is an applicant; 20--25\% of all applicants applied to Department~A; and 80\% of Department~A applicants were admitted.
Below, we show two possible 2D embeddings, where sets of applicants are represented by boxes.
The volumes of the boxes are shown in Table~\ref{tab:proportions}.
For example, on the left,
$\frac{\mathit{Vol}(\textit{DeptA} \sqcap \textit{Applicant})}{\mathit{Vol}(\textit{Applicant})} = \frac{\mathit{Vol}(\textit{DeptA})}{\mathit{Vol}(\textit{Applicant})} = \frac{8}{40} = 0.2$,
$\frac{\mathit{Vol}(\textit{Admitted} \sqcap \textit{DeptA})}{\mathit{Vol}(\textit{DeptA})} = \frac{6.4}{8} = 0.8$,
in line with the proportions stated in the conditionals at the top.
\end{example}

\begin{table}[h!]
    \centering
    \resizebox{0.45\textwidth}{!}{%
    \begin{tabular}{lcc}
    \toprule
    Concept & Vol. Left & Vol. Right \\
    \hline
    $Applicant$ & 40 & 60 \\
    $DeptA$ & 8 & 15 \\
    $Admitted$ & 17 & 50 \\
    $Admitted \sqcap DeptA$ & 6.4 & 12\\\bottomrule
    \end{tabular}
    }
    \caption{Volume of boxes in Figure \ref{fig:proportions}.}
    \label{tab:proportions}
\end{table}

We can now use the embeddings to estimate unknown proportions between concepts efficiently.
\begin{example}
To estimate the bounds for $(\textit{Admitted} \sqcap \textit{DeptA} \mid \textit{Applicant})[l_1, u_1]$, the proportion of all applicants who were admitted via Department~A, we take the minimum and maximum proportions observed across our two embeddings.
Hence, $l_1 = \min\!\bigl\{\tfrac{6.4}{40},\, \tfrac{12}{60}\bigr\} = 0.16$ and $u_1 = \max\!\bigl\{\tfrac{6.4}{40},\, \tfrac{12}{60}\bigr\} = 0.2$, resulting in the probability interval $[0.16, 0.2]$.
This is indeed the exact interval\footnote{The exact intervals can be computed by solving an (exponentially large) linear optimization problem.} as the embeddings happened to take the extreme values.

\end{example}
While our estimate was exact in the previous example,
the interval estimates can be too tight in other cases.

\begin{example} 
\label{exp_exact_interval}
\setcounter{footnote}{0}
For the overall admission rate $(\textit{Admitted} \mid \textit{Applicant})[l_2, u_2]$, we get
$l_2 = \min\!\bigl\{\tfrac{17}{40},\, \tfrac{50}{60}\bigr\} = 0.43$ and $u_2 = \max\!\bigl\{\tfrac{17}{40},\, \tfrac{50}{60}\bigr\} = 0.83$,
resulting in the interval estimate $[0.43, 0.83]$.
However, the exact interval is $[0.16, 0.96]$.
\end{example}
As we increase the number of embeddings, we increase the chance that one
of them yields an estimate close to the extreme values of the interval.
Therefore, we expect that increasing the number of embeddings
decreases the gap between the true and the estimated interval.

\section{A Normal Form for $\mathcal{SEL}$}\label{SEL norm}
\label{sec_normalization}

Embeddings of the classical DL $\mathcal{EL}$ \citep{KulmanovLYH19} exploit that every $\mathcal{EL}$ knowledge base can be transformed into an equivalent normal form that contains only 
GCIs of the following form \cite{baader2005pushing}: 
$$
A \sqsubseteq B,
A_1 \sqcap A_2 \sqsubseteq B,
A \sqsubseteq \exists r.B,
\exists r.A \sqsubseteq B,
$$
where $A$, $A_1, A_2, B$ are concept names or $\top$. Every $\mathcal{EL}$ TBox $\tbox$ can be transformed into a TBox $\tbox'$ in normal form with only a linear blowup of the size of the knowledge base \citep{baader2005pushing}.
We show the corresponding transformation rules in Table \ref{fig:transf_rules_el} in Appendix \ref{app:elnf}.
We can generalize the normal form to $\mathcal{SEL}$ as follows.

\begin{definition}[$\mathcal{SEL}$ normal form]
A $\mathcal{SEL}$ TBox $\mathcal{T}$ is in \emph{normal form} if for every conditional $(D \mid C)[l,u] \in \mathcal{T}$ 
\begin{itemize}
    \item $l = u = 1$ and
    $C \sqsubseteq D$ is in $\mathcal{EL}$ normal form or
    \item  $C, D \in \concepts$ are 
    concept names.
\end{itemize}
\end{definition}
That is, for a normalized conditional $(D \mid C)[l,u]$, either $C$ and $D$ are just concept names or the conditional is deterministic ($l=u=1$) and therefore just represents an $\mathcal{EL}$ GCI (c.f. Lemma
\ref{lemma_subs_and_eq}).
To normalize $\mathcal{SEL}$ TBoxes, we define a new rule that replaces complex concepts $C, D$ in non-deterministic conditionals with new concept names $A_1, A_2$:
\begin{align}
    SNF0 \qquad
    &(D \mid C)[l,u]\rightarrow  
    (A_2 \mid A_1)[l,u], \\
    &C \equiv A_1, D \equiv A_2,\nonumber
\end{align}
where $l<1$ and $C$ or $D$ are complex concepts. 
We can then normalize an $\mathcal{SEL}$ TBox as follows:
\begin{enumerate}
    \item Apply $SNF0$ to replace all complex concepts
    in conditionals with new concept names.
    \item Normalize the GCIs using the $\mathcal{EL}$ transformation rules from Table \ref{fig:transf_rules_el} in Appendix \ref{app:elnf}.
    \item Apply Lemma \ref{lemma_subs_and_eq} to transform the normalized GCIs into deterministic  conditionals.
\end{enumerate}
As we prove in Appendix \ref{app:proof}, the resulting $\mathcal{SEL}$ TBox is in $\mathcal{SEL}$ normal form, entailment-equivalent to the original TBox and its size is linear in the size of the original TBox.
\begin{proposition}
Let $\tbox$ be an $\mathcal{SEL}$ TBox and let $\tbox'$ denote the transformed TBox.
Then $\tbox'$ is in $\mathcal{SEL}$ normal form and its size is linearly bounded by the
    size of $\tbox$. Furthermore, for all $\mathcal{SEL}$ conditionals $(D \mid C)[l,u]$ built up over
    the concept names occurring in $\tbox$, we have
    $\tbox \models (D \mid C)[l,u]$ if and only if 
    $\tbox' \models (D \mid C)[l,u]$.
\end{proposition}
Note that the $\top$-concept does not have
a finite box representation. We must therefore
exclude it to avoid problems with the definition 
of proportions between (the volume of) boxes.
\begin{definition}
An $\mathcal{SEL}$ TBox $\tbox'$ 
is called \emph{safe} if for every probabilistic
conditional $(D \mid C)[l,u] \in \tbox'$, $\tbox' \not \models (C \equiv \top)$ and $\tbox' \not \models (D \equiv \top)$.
\end{definition}
Let us note that $\top$ can still occur in 
deterministic conditionals. However, probabilistic
facts $(D \mid \top)[l,u]$ that can be expressed in full 
$\mathcal{SEL}$ cannot be captured by our approach.

\section{Embedding and Approximation}
\label{encoding}
To perform approximate probabilistic reasoning, we construct geometric models for the safe normalized $\mathcal{SEL}$ ontologies $\mathcal{T}'$. In this section, we specify the mapping of concepts and roles to vector representations and formalize the loss terms that encode the probabilistic axioms. 




\textbf{Concept Representation.} Following \cite{XiongPTNS22, box2el}, we represent concepts by $n$-dimensional \emph{boxes}.
Formally, we represent the embedding of concepts by two functions $m_w, M_w$ that are parameterized by a learnable parameter vector $w$. In 2D, $m_w: \concepts \rightarrow \mathbb{R}^n$ maps concept names to the lower left corner, and  $M_w: \concepts \rightarrow \mathbb{R}^n$ maps them to the upper right corner of their box representation.
The \emph{box associated with $C \in \concepts$} is 
\begin{equation*}
\label{eq:volume}
    \Box_w(C) = \{x \in \mathbb{R}^n \mid m_w(C) \leq x \leq M_w(C) \},
\end{equation*}
where the inequality is defined elementwise-wise.
Its volume  is defined as:
\begin{equation*}
\label{box_volume}
     \operatorname{Vol}\left( \Box_w\left(C\right)\right) = \prod_{i=1}^n \max\left(0, M_w\left(C  \right)_i - m_w\left(C\right)_i\right).
\end{equation*}

\textbf{Role Representation.} Similar to \cite{BordesUGWY13, XiongPTNS22}, we associate every role name $r \in N_r$ with an affine transformation $T^r_w(x) = D^r_w x + b^r_w$, where $D^r_w$ is an $(n \times n)$ diagonal matrix with non-negative entries and $b^r_w \in \mathbb{R}^n$ is a vector. 
Applying $T^r_w$ to the box associated with a concept $C$ results in the box $T^r_w( \Box_w(C)) = \{T^r_w(x) \mid x \in \Box_w(C)\}$ with lower corner $T^r_w( m_w(C))$ and upper corner  $T^r_w( M_w(C))$.

When we compute an $n$-dimensional embedding, our parameter vector contains
$2n$ parameters for every concept name $C \in \concepts$ ($m_w(C)$ and
    $M_w(C)$), and
$2n$ parameters for every role name $r \in \roles$ ($b^r_w$ and diagonal entries of$D^r_w$).
Hence, the overall size of $w$ is $n\cdot \big(2 \cdot |\concepts| + 2 \cdot |\roles| \big)$. 


\textbf{Loss Function.} In order to compute the
parameter vector $w$ of our embedding, 
we minimize a loss function 
$\mathcal{L}(w)$. $\mathcal{L}(w)$ is composed
of multiple terms that correspond to the four axiom types 
that can occur in the normalized TBox.
We associate every axiom corresponding to a classical $\mathcal{EL}$ GCIs like in \cite{XiongPTNS22} by loss terms $\mathcal{L}_{NF1}(w), \mathcal{L}_{NF2}(w), \mathcal{L}_{NF3}(w), \mathcal{L}_{NF4}(w)$ corresponding to the four GCIs occurring in the classical normal form (Details can be found in Appendix \ref{app:boxel}).
We will introduce a new loss term to encode the meaning of probabilistic axioms.
Before we do so, let us note that every embedding can be seen as an $\mathcal{SEL}$ interpretation of the following type. 
\begin{definition}[Geometric Interpretation]
\label{def_el_interpretation}
A \emph{geometrical $\mathcal{SEL}$ interpretation (of dimension $n$)} $\interp = (\intdom, \intmap)$ is an $\mathcal{EL}$ interpretation where $\intdom = \mathbb{R}^n$.
\end{definition}

We associate the parameter vector $w$ of an $n$-dimensional $\mathcal{SEL}$ embedding with the $\mathcal{SEL}$ interpretation $\interp_w$ defined by
\begin{itemize}
    \item $C^{\interp_w} = \Box_w(C)$ for all $C \in \concepts$,
    \item $r^{\interp_w} = \{(a,b) \in \intdom \times \intdom \mid T^r_w(a) = b\}$ for all $r \in \roles$.
\end{itemize}
We can approximate the $\mathcal{SEL}$ semantics
with (infinite) geometric interpretations by replacing the cardinality of concepts with their volume. The satisfaction condition for the conditional $(D \mid C)[l,u]$ from equation \eqref{eq_sel_sat} then becomes
\begin{equation}
\label{eq_sel_box_sat}
    l \cdot \operatorname{Vol}(C^\interp) \leq \operatorname{Vol}(D^\interp \cap C^\interp)
\leq u \cdot \operatorname{Vol}(C^\interp).
\end{equation}
If one volume is infinite, we regard the condition
as violated.
We introduce one loss term for the upper and
one for the lower bound in \eqref{eq_sel_box_sat}
as follows:
\begin{align}
&\label{eq:loss_pcond_lower}
    \small
    \mathcal{L}^l_{(D \mid C)[l,u]}(w) =
    [l \cdot \operatorname{Vol}(C^\interp) -  \operatorname{Vol}(D^\interp \cap C^\interp)
      ]^+, \\
&\label{eq:loss_pcond_upper}
    \small
    \mathcal{L}^u_{(D \mid C)[l,u]}(w) = 
    [\operatorname{Vol}(D^\interp \cap C^\interp)
    - 
      u \cdot \operatorname{Vol}(C^\interp)]^+,
\end{align}
where $[x]^+ = \max \{0,x\}$.

\section{Soundness and Runtime}
\label{sec_correct_runtime}

We will now discuss some soundness and runtime guarantees of our approach. All proofs can be found in Appendix \ref{app:proof}.
As explained in the previous section, 
we compute the parameter vector
$w$ of the embedding by minimizing
\begin{equation}
\label{eq_tbox_loss}
    \mathcal{L}_\tbox(w) = \sum_{\alpha \in \tbox} \mathcal{L}_{\alpha}(w),
\end{equation}
where each axiom (conditional) $\alpha \in \tbox$ is associated with a loss term $\mathcal{L}_{\alpha}(w)$. 
The loss term for probabilistic axioms is composed of the lower and upper loss terms described in \eqref{eq:loss_pcond_lower}
and \eqref{eq:loss_pcond_upper}, that is,
\begin{equation}
\mathcal{L}_{(D \mid C)[l,u]}(w) = 
\mathcal{L}^l_{(D \mid C)[l,u]}(w) +
\mathcal{L}^u_{(D \mid C)[l,u]}(w).
\end{equation}

An embedding is called \emph{sound} if loss $\mathcal{L}_\tbox(w) = 0$ implies that
all axioms are satisfied by the geometric
interpretation $\interp_w$ associated with $w$ \citep{KulmanovLYH19}. As shown in \cite{XiongPTNS22}, for every axiom in $\mathcal{EL}$ normal form, $\mathcal{L}_{\alpha}(w) = 0$ implies that
$\interp_w$ satisfies $\alpha$. This implies soundness of the $\mathcal{EL}$-embedding in \cite{XiongPTNS22}. To prove  soundness of our $\mathcal{SEL}$-embedding, it suffices to show additionally that, for every probabilistic axiom $\alpha$, $\mathcal{L}_{\alpha}(w) = 0$ imply that $\interp_w$ satisfies $\alpha$ as well. 

\begin{lemma}
\label{lemma_sound_sel_loss_term}
If $\mathcal{L}_{(D \mid C)[l,u]}(w) = 0$, then $\interp_w \models (D \mid C)[l,u]$.
\end{lemma}

Together with the soundness results from \cite{XiongPTNS22}, we obtain the following soundness guarantee.
\begin{theorem}[Embedding Soundness]
\label{theorem_soundness}
If $\mathcal{L}_\tbox(w) = 0$, then  $\interp_w \models \tbox$.
\end{theorem}

The theorem also gives us a one-sided satisfiability test. If we can find an embedding with loss $0$, then $\tbox$ can be satisfied by a geometric $\mathcal{SEL}$ Interpretation. After computing the embedding, we want to use it to perform inference. As we prove in  Appendix \ref{app:proof}, this can be done efficiently.
\begin{theorem}[Runtime Guarantee]
\label{prop_inference_runtime}
For all (complex) concepts $C, D$ built up over $\concepts$
such that $\emptyset \neq C^{\interp_w}, D^{\interp_w} \neq \top^{\interp_w}$
, we can compute a $p \in [0,1]$
such that 
$\interp_w \models (D \mid C)[p]$ in
time $O((c+d)\cdot n)$, where
$c$ and $d$ are the sizes\footnote{Roughly speaking, the size of a concept corresponds to the number of constructors used to create it \citep{baader17book}.} of $C$ and $D$, respectively, 
and $n$ is
the embedding dimension.
\end{theorem}
If the embedding loss is $0$, our inference results are
sound in the following sense.
\begin{theorem}[Inference Soundness]
\label{thm_inf_soundness}
 If $\mathcal{L}_\tbox(w) = 0$ and $\tbox \models (D \mid C)[l, u]$,
then $\interp_w \models (D \mid C)[p]$ implies $p \in [l,u]$.   
\end{theorem}
Intuitively, the result guarantees that our computations are exact when the 
statistical proportions can be embedded perfectly.
If there is an embedding error, this error will propagate through our
probability estimates and we will evaluate the 
approximation quality in such cases empirically in Section \ref{sec:exp}.
To efficiently generate valid consequences of a given TBox,
we derive a probabilistic generalization of
the classical \emph{Modus Ponens} for $\mathcal{SEL}$ 
in Appendix \ref{app:proof}.
\begin{proposition}[Probabilistic Modus Ponens (PMP)]\label{prop:pmp}
For all $\mathcal{EL}$ concepts $C, D, E$ (atomic or complex)
and all $\mathcal{SEL}$ interpretations $\interp$,
if $\tbox \models (D \mid C)[l_1, u_1]$ and
$\tbox \models (E \mid C \sqcap D)[l_2, u_2]$,
then $\tbox \models (E \mid C)[l_3, u_3]$,
where $l_3 = l_1 \cdot l_2$
and $u_3 = \min \{1, u_1 \cdot u_2 + 1 - l_1\}$.
\end{proposition}

\section{Experiments}\label{sec:exp}


Before presenting the experimental results,
we will describe our implementation, 
the procedure for constructing test ontologies
and our evaluation protocol.

\subsection{Implementation}
\label{sec_implementation}


We implemented a first prototype which supports conditionals of the following types:
\begin{align}
\label{eq_supported_conditionals}
(B|A)[p],\ &(B|A_1\sqcap A_2)[p], (B|\exists r.A)[p], (\exists r.B|A)[p],
\end{align}
where $A, A_1, A_2, B \in \concepts$ are concept names.
As their structure 
corresponds to the four axioms occuring in $\mathcal{EL}$ normal form, we will refer to them as PNF1 - PNF4.

\cite{XiongPTNS22} added a  \emph{location regularizer} $\mathcal{L}_{loc}(w)$
to the loss function that is defined as 
\begin{equation}
 \sum_{C \in \concepts} \sum_{i=1}^{n}
 \  [M_w(C)_i - \beta+\epsilon]^+ + [-m_w(C)_i - \epsilon]^+,
\end{equation}
where $\beta$ is a hyperparameter that bounds the upper and lower corner of boxes. 
Intuitively, $\mathcal{L}_{loc}(w)$ encourages that concepts
are embedded in the hypercube $[0,\beta]^n$.
As we show in the ablation study in Table \ref{tab:abl_reg},
just regularizing the volume of boxes can yield better results.
We therefore also consider the following \emph{volume regularizer}:
\begin{equation}
    \mathcal{L}_{vol}(w) = \sum_{i=1}^m \ [\beta^n - Vol(C_{i})-\epsilon]^+
\end{equation}
Even though $\mathcal{L}_{loc}$  and $\mathcal{L}_{vol}$ have some redundancy,
we found that their combination can work better than each term individually. 
The loss function that we minimize is therefore
\begin{equation}
   \mathcal{L}(w) = \mathcal{L}_\tbox(w) + \mathcal{L}_{loc}(w) + \mathcal{L}_{vol}(w). 
\end{equation}

We performed hyperparameter search for 
embedding dimensions $n = [8, 16, 32, 64, 128]$, 
side lengths $\beta = [1,10]$ and
learning rates $lr = [0.001, 0.01, 0.05, 0.1]$
for Adam \citep{KingBa15}. Table \ref{tab:hyper_param} summarizes the best hyperparameters. 
Our experiments ran on a Linux machine with  a 40GB NVIDIA A100 SXM4 GPU. 
We provide additional details in Appendix \ref{app:impl}.

\subsection{Dataset Construction}\label{datat_generation}
To the best of our knowledge, there are currently no $\mathcal{SEL}$ ontologies available for evaluating our method. Therefore, we derived three datasets—\emph{Country}, \emph{Hybrid}, and \emph{Person}— from the high-quality real-world knowledge base YAGO3 \citep{mahdisoltani2014yago3}. These ontologies were created by computing statistics for conditionals of types PNF1 to PNF4 (c.f., Equation \eqref{eq_supported_conditionals}).

Specifically, we first select a limited number of concepts and role names of interest. We then systematically generate all possible conditionals of types PNF1 to PNF4 that can be created from the selected concepts and role names, assigning probabilities based on statistical proportions. For example, to compute the probability $p$ of the probabilistic conditional $(\exists hasChild\mid Person)[p]$, we count the number of persons who have children and divide it by the total number of persons in YAGO3. More details about the dataset construction process can be found in Appendix \ref{app:data generation}. 

It is important to note that we consider only a subset of the theoretically possible conditionals due to the following 
constraints:
\begin{itemize}
    \item \textbf{Undefined Probabilities}: Conditionals with undefined probabilities are excluded. For instance, $(B|A_1\sqcap A_2)$ is undefined if $A_1$ and $A_2$ are disjoint because the probability of the condition is $0$.
    \item \textbf{Redundancy}: We removed redundant axioms. For example, if $A_1$ and $A_2$ are disjoint, conditionals stating that their subconcepts are disjoint are redundant.
\end{itemize}

Table \ref{tab:dataset} presents statistical information about the ontologies after the filtering process.

\begin{table}[t]
\centering
\resizebox{0.473\textwidth}{!}{
\begin{tabular}{cccc}
\toprule
                 & \#Concepts & \#Role Names & \#T-Box Axioms (the ratio of PNF1-4) \\ \midrule
Person  & 135         & 6        & 325,600 (0.21/0.31/0.24/0.24)  \\
Country & 30         & 1        & 26,974 (0.24/0.5/0.13/0.13)  \\
Hybrid  & 12         & 18        & 7,492 (0.02/0.22/0.38/0.38)  \\ \bottomrule
\end{tabular}
}
\caption{Statistics of our benchmark ontologies.}
\label{tab:dataset}
\end{table}

\subsection{Experimental Setup}
Our reasoning task involves $\mathcal{EL}$ concepts $C, D, E$ (atomic or complex). Given premises $(D \mid C)[l_1, u_1]$, $(E \mid C \sqcap D)[l_2, u_2]$, and a query $(E \mid C)$, our goal is to infer the associated probability interval $[l_3, u_3]$. As we illustrated in the examples in Section \ref{sec:example}, if our embedding approximately captures the proportions stated in premise conditionals, we should be able to approximate the probability interval by calculating proportions between boxes in the embedding space. Theorem
\ref{thm_inf_soundness} implies that, whenever the embedding loss is $0$, the probability interval cannot be too large. However, it can be too small as illustrated in Example \ref{exp_exact_interval}. Furthermore, the embedding error will typically be non-zero, and so the interval can also be too large.  

\textbf{Data Split.} 
We split axioms in $\mathcal{SEL}$ ontologies into two subsets: a \emph{learning set} for training embeddings and a \emph{query set} for evaluating approximation quality. In our experiment, we select 30\% of axioms from the ontology as the query set for evaluation.
Importantly, the split is non-random. For each query $(E \mid C)$, we remove the corresponding conditional $(E \mid C)[l_3, u_3]$ from the ontology and ensure that the premises $(D \mid C)[l_1, u_1]$ and $(E \mid C \sqcap D)[l_2, u_2]$ are present in the ontology to allow non-trivial inferences.

\textbf{PMP Evaluation.} 
Let $q$ denote the number of queries. By construction, each query has corresponding premise conditionals in the ontology, allowing us to apply PMP (Proposition \ref{prop:pmp}) to infer a probability interval $[l, u]$. To approximate this interval, we compute $N$ embeddings with different random seeds, yielding $N$ point estimates ${p_1, \dots, p_N}$ (by computing proportions in the embedding) for each query. The lower and upper bounds are estimated as $\overline{l} = \min \{p_1, \dots, p_N\}$ and $\overline{u} = \max \{p_1, \dots, p_N\}$, respectively. In our experiments, we typically generate five query sets by varying the data split and set $N = 60$.

A detailed description of the query set construction and the pseudocode for PMP evaluation are provided in 
Appendix \ref{app:PMP}.

\subsection{Evaluation Protocol}

\textbf{Metrics.} 
We evaluate the \emph{embedding quality} using \emph{Mean Absolute Error} (MAE), which measures the average absolute difference between the probabilities in the knowledge base and the corresponding proportions in the embedding. 

In order to evaluate the \emph{inference quality} of our approximation
$[\overline{l},\overline{u}]$, we consider three different metrics. 
\begin{itemize}
    \item \emph{Soundness Accuracy} (SA): This metric measures the percentage of queries for which the approximate interval fell into the true interval.
    \begin{equation}
        SA := \frac{1}{q}\sum_{i=1}^q \textbf{1}_{
        [\overline{l}_i,\overline{u}_i] \subseteq [l_i,u_i]
        },
    \end{equation}
    where $\textbf{1}_{S \subseteq T}$ is $1$ if $S \subseteq T$ and $0$ otherwise.
    \item \emph{Soundness Error} (SE): This metric quantifies the extent to which the estimated interval $[\overline{l}, \overline{u}]$ extends beyond (overestimates) the true interval $[l, u]$.
    \begin{equation}\label{sound_error}
       SE := \frac{1}{q}\sum_{i=1}^q \left( [l_i - \overline{l}_i]^+ + [\overline{u}_i - u_i]^+ \right).
    \end{equation}
    \item \emph{Approximation Gap} (AG): This metric captures the absolute difference between the estimated and true lower/upper bounds.
    \begin{equation}
     AG :=   \frac{1}{q}\sum_{i=1}^q\left(|l_i - \overline{l}_i| + |u_i - \overline{u}_i|\right).
    \end{equation}
\end{itemize}

\textbf{Baselines.} 
To the best of our knowledge, there are currently no existing implementations for statistical reasoning in $\mathcal{SEL}$ that can be used for direct comparison (see the discussion in Section \ref{sec:related_work}). To demonstrate that our approach gives non-trivial solutions,
we consider several simple baselines to serve as points of reference for evaluating the performance of our approach:
\begin{itemize}
    \item \emph{Fixed Estimator}: An interval estimator that always outputs $[0,1]$ as the predicted interval. This estimator would perform (almost) perfectly for SA, SE and AG if all true intervals were (close to) $[0,1]$.
    \item \emph{Random Estimator}: An interval estimator that randomly samples $\bar{l}, \bar{u}$ from $[0,1]$, ensuring $\bar{l} \leq \bar{u}$, as the predicted interval. The motivation is to demonstrate that our method works significantly better than a random guess.
    \item \emph{KDE Estimator}: This method estimates the probability density function (PDF) of the intervals in the learning set using a Kernel Density Estimation (KDE) method \citep{davis2011remarks}. It then samples an interval from the estimated PDF as the predicted interval. Further details are provided in Appendix \ref{app:kde_estimator}.
\end{itemize}

Note that existing methods (e.g. ELEm, EMEL$^{++}$, BoxEL and Box2EL) only handle classical $\mathcal{EL}$ knowledge bases and not $\mathcal{SEL}$ knowledge bases. Therefore, we cannot consider them as baselines for inference in $\mathcal{SEL}$. However, we perform ablation studies on alternative design choices inspired by these methods in Section \ref{sec:ablation}.

\subsection{Experimental Results and Analysis}
We now present and discuss the results of our experiments, focusing on the following questions: \textbf{RQ1}: How effectively does our approach approximate statistical reasoning in $\mathcal{SEL}$? \textbf{RQ2}: What is the relationship between embedding quality and the quality of approximate inference? \textbf{RQ3}: How efficient is the inference process of our approach?

\begin{table}[t]
\centering
\resizebox{0.5\textwidth}{!}{%
{\small
\begin{tabular}{clccc}
\toprule[1.2pt]
Dataset & \multicolumn{1}{c}{Method} & SA $\uparrow$ & SE $\downarrow$ & AG $\downarrow$ \\\midrule
 & Fixed Estimator & 4.50\% & 0.301 & 0.301 \\
 & Random Estimator & {\color[HTML]{000000} 52.70\%} & 0.262 & 0.572 \\
 & KDE Estimator & 9.19\% & 0.241 & 0.275 \\
\multirow{-4}{*}{Person} & BoxSEL (ours) & \textbf{89.30\%} & \textbf{0.013} & \textbf{0.233} \\\midrule
 & Fixed Estimator & 29.10\% & 0.105 & \textbf{0.105} \\
 & Random Estimator & {\color[HTML]{000000} 61.30\%} & 0.048 & 0.603 \\
 & KDE Estimator & 3.10\% & 0.086 & {\color[HTML]{000000} 0.329} \\
\multirow{-4}{*}{Country} & BoxSEL (ours) & \textbf{95.60\%} & \textbf{0.004} & 0.241 \\\midrule
 & Fixed Estimator & 1.5\% & 0.734 & 0.734 \\
 & Random Estimator & 27.8\% & 0.537 & 0.533 \\
 & KDE Estimator & 1.1\% & 0.661 & 0.353 \\
\multirow{-4}{*}{Hybrid} & BoxSEL (ours) & \textbf{89.70\%} & \textbf{0.020} & \textbf{0.293}\\\bottomrule[1.2pt]
\end{tabular}%
}
}
\caption{Approximation quality of baselines and BoxSEL.}
\label{tab:main}
\end{table}

\begin{figure}[t]
\centering
\includegraphics[width=.38\textwidth]{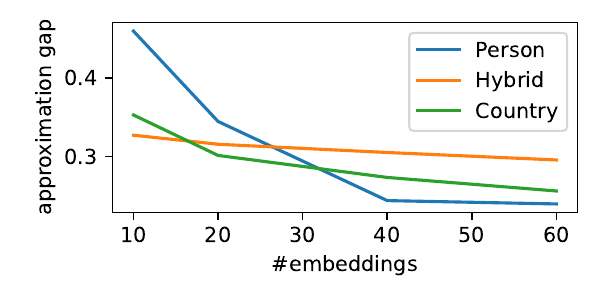}
\caption{Approximation gap for approximation based on 10, 20, 40 and 60 embeddings.}\label{fig:app_gap}
\end{figure}

\begin{figure}[t]
\centering
\includegraphics[width=.48\textwidth]{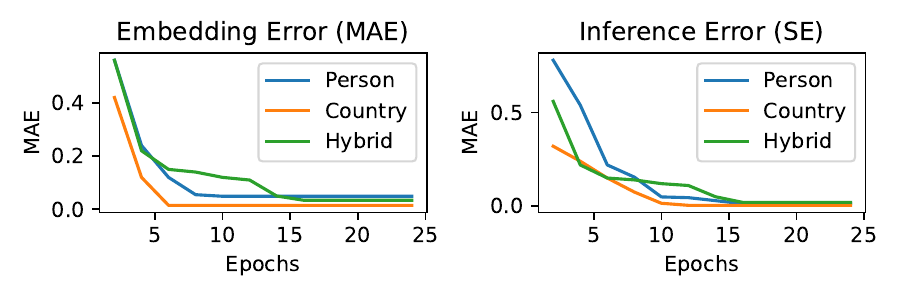}
\caption{Relationship between embedding and inference error.}\label{fig:rel_EmdInf}
\end{figure}

\begin{table}[t]
\centering
\resizebox{.43\textwidth}{!}{
\begin{tabular}{cccccc}
\toprule[1.2pt]
 & Embedding (s/epoch)
 & Inference (ms/query)\\ \midrule
Person  & 117.66 
& 0.09\\
Country & 8.96 
& 0.23\\
Hybrid  & 12.15 
& 0.41\\ \bottomrule[1.2pt]
\end{tabular}
}
\caption{Evaluation of runtime of computing the embedding and performing inference on the computed embedding.}
\label{tab:runtime}
\end{table}

\subsubsection{Approximation Quality}
Table \ref{tab:main} presents the approximation quality for baselines and our method, BoxSEL, across three test $\mathcal{SEL}$ ontologies: Person, Country, and Hybrid.
Our approach significantly outperforms the baselines across $SA$ and $SE$.
Specifically, a large proportion of the approximations of our approach is sound in the sense that the estimated intervals are contained in the true intervals (SA close to 100\%) and the soundness error is relatively small ($2 \%$ absolute error or less).
The approximation gap for BoxSEL is between 0.2 and 0.3. Let us note that the approximation gap for the fixed estimator is about 0.3 for the Person and 0.1 for the Country dataset. This shows that the ground truth intervals in these benchmarks are relatively large (the average width is 0.7 for Person and 0.9 for Country). We note that the approximation gap of our approach is not very sensitive with respect to the width of the ground truth interval. While
we can see an increase in precision (SA and SE) for benchmarks with smaller intervals, it remains relatively large (above 89 \%) for all benchmarks
independent of the width of the ground truth intervals.

Figure \ref{fig:app_gap} illustrates how the approximation gap decreases as the number of embeddings used increases. This confirms our intuition since adding more samples increases the chance of picking a sample close to the extreme values of the interval.




\subsubsection{Inference vs Embedding Error}
We expect that the inference error 
(how well do we approximate the query interval?)
decreases with the
embedding error (how well could the axioms in the ontology be represented by embeddings?) because errors in the embedding of proportions can cause errors
in the derived estimates. In Figure \ref{fig:rel_EmdInf}, we illustrate this relationship
by plotting embedding and inference
error against the number of epochs used for computing the 
embeddings. We can see a clear relationship between inference and embedding error.
This is a useful property in practice because the embedding error is known
after computing the embedding. If the embedding error is low,
we expect good approximation results. If the embedding error is large, we should be more skeptical about the inferred probability intervals and
invest more time into finding better embeddings if a higher accuracy is needed. 

\subsubsection{Runtime}
Table \ref{tab:runtime} shows runtime results for computing embeddings (per epoch) and performing inference (per query) on the embedding. Note that each embedding has to be computed only once to transform an ontology into an embedding. Afterwards,
the embedding can be used to answer an arbitrary number of queries over the ontology.
Table \ref{tab:runtime} shows that
even a training epoch for the large \emph{Person}
dataset can be performed in less than two minutes on average. 
Since embeddings are computed by gradient descent variants,
the runtime per epoch depends linearly on the size of the embedding vector and the number of axioms. As we explained at the end of Section \ref{encoding}, the embedding vector grows linearly with respect to the individual, concept, and role names. So, overall, the  runtime for computing embeddings is linear with respect
to the size of the knowledge base and the number of epochs. 
For the experiments in Table \ref{tab:main},
computing one embedding for the 
person/country/hybrid dataset took 39/3/6 minutes on average.
For all datasets, inference is a matter of milliseconds as we can expect from the linear runtime guarantees stated in Theorem \ref{prop_inference_runtime}.

\subsection{Ablation Study}\label{sec:ablation}
We conduct three ablation studies to investigate the impact of different concept and role representation, as well as the effect of regularization, on the approximation quality.

\textbf{Impact of Concept Representation.}
Inspired by ELEm and EMEL$^{++}$, we explored an alternative representation of concepts as \emph{n-balls}, while keeping all other components of our method unchanged. 
The results, presented in Table \ref{tab:abl_concept}, show that representing concepts as boxes consistently outperforms the n-balls representation across all metrics. These findings emphasize the advantages of using boxes for concept representation, both theoretically and empirically.

\textbf{Impact of Role Representation.} 
There are several alternative methods for representing roles in the literature. ELEm and EMEL$^{++}$ use translation, BoxEL employs affine transformation, and Box2EL utilizes dual boxes to represent roles. We compare these three approaches in Table \ref{tab:abl_role}.
Our findings indicate that affine transformation achieves the highest performance in the Person and Country ontologies and the dual boxes representation slightly outperforms affine transformation in the Hybrid ontology. 
However, this improvement comes at the cost of increased model complexity.
Dual boxes require $4n\cdot |\roles|+n\cdot |\concepts|$ parameters to represent roles \citep{box2el}, which is more than twice the parameters required by the affine transformation ($2n\cdot |\roles|$).
Additionally, the dual boxes approach shows reduced performance in the Person and Country ontologies.

\begin{table}[t]
\centering
{\small
\resizebox{0.5\textwidth}{!}{%
\begin{tabular}{ccccc}
\toprule[1.2pt]
Dataset & Role Representation & SA $\uparrow$ & SE $\downarrow$ & AG $\downarrow$ \\\midrule
 & translation & 87.3\% & 0.017 & \textbf{0.232} \\
 & affine transformation$^*$ & \textbf{89.3\%} & \textbf{0.013} & 0.233 \\
 \multirow{-3}{*}{Person} & dual boxes & 85.7\% & 0.109 & 0.279 \\\midrule
 & translation & 94.7\% & 0.005 & 0.245 \\
 & affine transformation$^*$ & \textbf{95.6\%} & \textbf{0.004} & \textbf{0.241} \\
 \multirow{-3}{*}{Country} & dual boxes & 93.2\% & 0.013 & 0.261 \\\midrule
 & translation & 79.8\% & 0.039 & 0.293 \\
 & affine transformation$^*$ & 89.7\% & \textbf{0.020} & 0.293 \\
 \multirow{-3}{*}{Hybrid} & dual boxes & \textbf{89.8\%} & 0.031 & \textbf{0.287} \\\bottomrule[1.2pt]
\end{tabular}%
}
}
\caption{Ablation study of the role representation. The design choice of our method is marked with an asterisk (*).}
\label{tab:abl_role}
\end{table}

\textbf{Impact of Regularization.} We investigate the impact of the regularization terms ($\mathcal{L}_{loc}$ and $\mathcal{L}_{vol}$). Table \ref{tab:abl_reg} shows that volume regularization significantly improves performance. Location regularization further enhances performance and reduces the number of epochs required for convergence.

\section{Discussion and Future Work}

We showed how embeddings can be used to approximate probabilistic inference in $\mathcal{SEL}$ by extending the $\mathcal{EL}$-embedding BoxEL from \cite{XiongPTNS22}. 
As shown in Section~\ref{sec_correct_runtime}, the resulting embeddings support point estimates for $\mathcal{SEL}$ queries in linear time, with guaranteed soundness when the embedding loss is zero. When the loss is non-zero, the inference error increases, but our experiments show a clear proportional relationship between the two, making the known embedding error a practical quality indicator. To approximate full probability intervals, we compute multiple embeddings and combine the individual point estimates. In our experiments, this procedure rarely overestimated the true intervals, and approximation quality improved steadily as the number of embeddings increased. In practice, $\mathcal{SEL}$ ontologies can be constructed fully automatically from arbitrary knowledge graphs by applying rule mining methods \citep{OmranWW18,MeilickeCRS19,GuimarPS21} and using the confidence values (which are usually statistical proportions) of rules for the probabilities.

Our inference soundness guarantee (Theorem~\ref{thm_inf_soundness}) also sheds light on BoxEL from \citet{XiongPTNS22}: if all conditionals in an $\mathcal{SEL}$ ontology are deterministic, it reduces to an $\mathcal{EL}$ ontology (Lemma~1), and our $\mathcal{SEL}$ embedding becomes equivalent to the BoxEL embedding. Our soundness theorem thus guarantees that inference under BoxEL embeddings is sound as well when the loss is zero, and our empirical results suggest that the approximation error remains small when the loss is small.

\section{Acknowledgements}
The authors thank the International Max Planck Research
School for Intelligent Systems (IMPRS-IS) for supporting Yuqicheng Zhu. This work was partially supported by the Deutsche Forschungsgemeinschaft (DFG, German Research Foundation) - SFB 1574 - Project number 471687386, and by the EU Project SMARTY (GA 101140087).

\bibliography{uai2026-template}

\newpage

\onecolumn

\appendix
\section{EL Transformation Rules}\label{app:elnf}
\begin{table}[h]
    \centering
    \resizebox{0.45\textwidth}{!}{%
    \begin{tabular}{llll}
        $NF0$ & $\hat{D} \sqsubseteq \hat{E}$ & 
        $\rightarrow$ & 
        $\hat{D} \sqsubseteq A$,
        $A \sqsubseteq \hat{E}$.
        \\
        $NF1_r$ & $C \sqcap \hat{D} \sqsubseteq B$ 
        &$\rightarrow$ 
        &
        $\hat{D} \sqsubseteq A$,
        $C \sqcap A \sqsubseteq B$.\\
        $NF1_l$ &  $\hat{D} \sqcap C \sqsubseteq B$ 
        &$\rightarrow$ 
        &
        $\hat{D} \sqsubseteq A$,
        $A \sqcap C \sqsubseteq B$.\\
        $NF2$ & $\exists r.\hat{D} \sqsubseteq B$ 
        &$\rightarrow$ 
        & 
        $\hat{D} \sqsubseteq A$,
        $\exists r.A \sqsubseteq B$.\\
        $NF3$ & $B \sqsubseteq \exists r.\hat{D}$ 
        &$\rightarrow$ 
        & 
        $A \sqsubseteq \hat{D}$,
        $B \sqsubseteq \exists r.A$.\\
        $NF4$ & $B \sqsubseteq D \sqcap E$ 
        &$\rightarrow$ 
        &
        $B \sqcap D$,
        $B \sqcap E$.
    \end{tabular}
    }
    \caption{Transformation rules for
    normalizing $\mathcal{EL}$ TBoxes 
    from \protect\cite{baader17book}. 
    $C, D, E$
    are arbitrary $\mathcal{EL}$
    concepts, $\hat{D}, \hat{E}$
    are non-atomic concepts,
    $B$ is a concept name and $A$ is a new concept name.}
    \label{fig:transf_rules_el}
\end{table}

\section{Resolving Simpson's Paradox with $\mathcal{SEL}$}\label{app:simpson}

We illustrate how $\mathcal{SEL}$ can detect Simpson's paradox using a simplified version of the UC Berkeley admissions scenario from Section~\ref{sec:example}. Consider two departments, A and B, with the following conditionals about female applicants:
\begin{align*}
    (\textit{Admitted} \mid \textit{DeptA} \sqcap \textit{Woman})&[0.80], \\
    (\textit{Admitted} \mid \textit{DeptB} \sqcap \textit{Woman})&[0.90], \\
    (\textit{DeptA} \mid \textit{Woman})&[0.90], \\
    (\textit{DeptB} \mid \textit{Woman})&[0.10],
\end{align*}
and the corresponding conditionals for male applicants:
\begin{align*}
    (\textit{Admitted} \mid \textit{DeptA} \sqcap \textit{Man})&[0.75], \\
    (\textit{Admitted} \mid \textit{DeptB} \sqcap \textit{Man})&[0.85], \\
    (\textit{DeptA} \mid \textit{Man})&[0.10], \\
    (\textit{DeptB} \mid \textit{Man})&[0.90].
\end{align*}

Women are admitted at higher rates than men in \emph{both} departments (80\% vs.\ 75\% in~A, 90\% vs.\ 85\% in~B). However, 90\% of female applicants apply to Department~A, which has the lower admission rate, while 90\% of male applicants apply to Department~B.

Given these conditionals, $\mathcal{SEL}$ can derive the overall admission rates by querying $(\textit{Admitted} \mid \textit{Woman})$ and $(\textit{Admitted} \mid \textit{Man})$. Since departments A and B partition the applicants of each gender, the law of total probability gives:
\begin{align*}
    P(\textit{Admitted} \mid \textit{Woman}) &= 0.80 \times 0.90 + 0.90 \times 0.10 = 0.81, \\
    P(\textit{Admitted} \mid \textit{Man}) &= 0.75 \times 0.10 + 0.85 \times 0.90 = 0.84.
\end{align*}
The aggregate male admission rate (0.84) exceeds the aggregate female rate (0.81), even though women are admitted at higher rates in every individual department. This is Simpson's paradox: the uneven distribution of applicants across departments reverses the comparison at the aggregate level.

$\mathcal{SEL}$ makes this reversal explicit. By encoding department-level admission rates and applicant distributions as conditionals, and then querying the aggregate rates, the paradox becomes a direct consequence of the derived bounds rather than a hidden artifact of aggregation.

\section{Proofs of Technical Results}\label{app:proof}
\setcounter{proposition}{0}
\setcounter{lemma}{1}
\setcounter{theorem}{0}

To improve comprehensibility of the proof, we rewrite Proposition 1 slightly by enumerating the main claims.
\begin{proposition}
Let $\tbox$ be an $\mathcal{SEL}$ TBox
and let $\tbox'$ denote the TBox resulting from our 
transformation procedure.
\begin{enumerate}
    \item $\tbox'$ is in $\mathcal{SEL}$ normal form.
    \item For all $\mathcal{SEL}$ conditionals $(D \mid C)[l,u]$ built up over
    the concept names occurring in $\tbox$, we have
    $\tbox \models (D \mid C)[l,u]$ if and only if 
    $\tbox' \models (D \mid C)[l,u]$.
    \item The size of $\tbox'$ is linearly bounded by the
    size of $\tbox$.
\end{enumerate}
\end{proposition}
\begin{proof}
1. After step 1, all remaining conditionals are non-deterministic
and contain only concept names. Therefore, they satisfy condition 2 of the normal form. In step 2 and 3, the equivalences are translated into
GCIs and normalized. Step 4 translates the normalized GCIs
into deterministic conditionals. Since the GCIs were in 
$\mathcal{EL}$ normal form, the resulting deterministic conditionals satisfy condition 1 of the normal form. Hence, $\tbox'$ is in $\mathcal{SEL}$ normal form.

2. First assume $\tbox \models (D \mid C)[l,u]$  and
let $\interp'$ be an interpretation that satisfies
$\tbox'$. Then, by construction of $\tbox'$,
the interpretation $\interp$ obtained from $\interp'$
by restricting to the concept names occurring in $\tbox$
also satisfies $\tbox$. Hence, by assumption, $\interp \models
(D \mid C)[l,u]$. Since $\interp$ and $\interp'$ interpret
the concept names occurring in $\tbox$ equally, 
we also have $\interp' \models
(D \mid C)[l,u]$.

Conversely, assume $\tbox' \models (D \mid C)[l,u]$  and
let $\interp$ be an interpretation that satisfies
$\tbox$. Extend $\interp$ to an interpretation $\interp'$
of $\tbox'$ as follows: for every new concept name $A$
occurring in $\tbox'$ but not in $\tbox$, there must be 
a GCI $A \sqsubseteq C$. In particular, we can order the new
concept names such that $C$ contains only concept names that
have already been interpreted (just follow the order that the
normalization algorithm used when it introduced new concept names). We then let $A^{\interp'} = C^{\interp'}$. Then
$\interp'$ satisfies $\tbox'$ as well and, by assumption,
$\interp'$ satisfies $(D \mid C)[l,u]$. Since, $\interp'$
extends $\interp$, $\interp$ satisfies $(D \mid C)[l,u]$ as well. 

3. After step 1, every conditional is replaced by at most one
conditional and two equivalences. The length of each of these
is bounded by the length of the original conditional. Hence,
the size remains linear after step 1. In step 2, each 
equivalence is replaced by 2 GCIs of the same length. 
Hence, for every conditional, we introduce at most 4 GCIs 
whose length is at most the length of the original conditional.
We then apply the $\mathcal{EL}$ normalization rules that
guarantee that the size of the resulting GCIs is linear in
the size of the original GCIs. Hence, it is also linear
in the size of the original conditionals (see Lemma 6.2 in \cite{baader17book}). Hence, after translating the 
GCIs into deterministic conditionals in step 4, the size of 
$\tbox'$ is linear in the size of $\tbox$.
\end{proof}



\begin{lemma}
If $\mathcal{L}_{(D \mid C)[l,u]}(w) = 0$, then $\interp_w \models (D \mid C)[l,u]$.
\end{lemma}

\begin{proof}
Since $\mathcal{L}^l_{(D \mid C)[l,u]}(w)$ and
$\mathcal{L}^u_{(D \mid C)[l,u]}(w)$ are non-negative by definition,
$\mathcal{L}_{(D \mid C)[l,u]}(w) = 0$ implies that 
both terms are equal to 0. Hence, $l \cdot \operatorname{Vol}(C^\interp) \leq \operatorname{Vol}(D^\interp \cap C^\interp)$, $\operatorname{Vol}(D^\interp \cap C^\interp)\leq u \cdot \operatorname{Vol}(C^\interp)$ and $\interp_w \models (D \mid C)[l,u]$.
 \end{proof}

\begin{theorem}[Embedding Soundness]
If $\mathcal{L}_\tbox(w) = 0$, then  $\interp_w \models \tbox$.
\end{theorem}
\begin{proof}
All loss terms in $\mathcal{L}_\tbox(w) = \sum_{\alpha \in \tbox} \mathcal{L}_{\alpha}(w)$ are non-negative by definition.
Hence, $\mathcal{L}_\tbox(w) = 0$ implies $\mathcal{L}_{\alpha}(w) = 0$ for all $\alpha \in \tbox$.
$\mathcal{L}_{\alpha}(w) = 0$ together with
Proposition 1-5 from \cite{XiongPTNS22} (for deterministic axioms) along with
Lemma $\ref{lemma_sound_sel_loss_term}$ (for probabilistic axioms) 
imply that $\interp_w \models \alpha$ for all $\alpha \in \tbox$.
Hence,  $\interp_w \models \tbox$.
 \end{proof}

\begin{theorem}[Runtime Guarantee]
\label{prop_inference_runtime}
For all (complex) concepts $C, D$ built up over $\concepts$
such that $\emptyset \neq C^{\interp_w}, D^{\interp_w} \neq \top^{\interp_w}$
, we can compute a $p \in [0,1]$
such that 
$\interp_w \models (D \mid C)[p]$ in
time $O((c+d)\cdot n)$, where
$c$ and $d$ is the size of $C$ and $D$, respectively, 
and $n$ is
the embedding dimension.
\end{theorem}
\begin{proof}
In order to compute $p$, we have to compute     
$\operatorname{Vol}(D^\interp \cap C^\interp)$ and
$\operatorname{Vol}(C^\interp)$
and can then let $p = \frac{\operatorname{Vol}(D^\interp \cap C^\interp)}{\operatorname{Vol}(C^\interp)}$. We will now explain how we 
can compute the volume.

If $C, D$ are concept names, $\operatorname{Vol}(C^\interp)$
can be directly computed from \eqref{eq:volume},
which takes time $O(n)$.
To compute $\operatorname{Vol}(D^\interp \cap C^\interp)$,
we first have to compute $\Box_w\left(C^\interp \cap D^\interp \right)$.
We have
$m_w(C^\interp \cap D^\interp) = \max \{m_w(C^\interp), 
m_w(D^\interp)\}$
for the lower corner
and 
$M_w(C^\interp \cap D^\interp) = \min \{M_w(C^\interp), 
M_w(D^\interp)\}$
for the upper corner, where minimum and maximum are taken componentwise.
If one component of $M_w(C^\interp \cap D^\interp) - m_w(C^\interp \cap D^\interp)$ is negative, the intersection is empty and the volume is $0$.
Otherwise, we can again apply equation  \eqref{eq:volume} to compute
the volume in time $O(n)$.

If $C$ or $D$ are not concept names, we decompose them recursively.
We demonstrate this for $C$. 

If $C = C_1 \sqcap C_2$, then we can 
associate $C$ with a box as in the previous case in time $O(n)$
and continue with this box. 

$C = \exists r.C_1$,
the box associated with $C$ is defined by
$m_w(C) = A m_w(C_1)$
and $M_w(C) = A M_w(C_1)$
, where $A$ is the inverse matrix of $T^r_w$. Since 
$T^r_w$ is a diagonal matrix, the inverse matrix is obtained by simply inverting the
diagonal entries ($x \mapsto \frac{1}{x}$).
Since $A$ is a diagonal matrix as well, we only have to multiply the diagonal 
entries of $A$ by the entries in the vector, so that the overall runtime
is again $O(n)$. We then continue again with the box representation of $C$.

Each decomposition step can be performed in time $O(n)$ and the number of
decomposition steps is bounded by the size of $C$ and $D$.
Therefore, the overall runtime is $O((c+d)\cdot n)$.
\end{proof}

\begin{theorem}[Inference Soundness]
 If $\mathcal{L}_\tbox(w) = 0$ and $\tbox \models (D \mid C)[l, u]$,
then $\interp_w \models (D \mid C)[p]$ implies $p \in [l,u]$.   
\end{theorem}
\begin{proof}
If $\mathcal{L}_\tbox(w) = 0$, our soundness guarantee implies that 
$\interp \models \tbox$. Since $\tbox \models (D \mid C)[l, u]$,
every model of $\tbox$ respects the bounds $l$ and $u$.
Hence, we must have $p \in [l,u]$.
\qed \end{proof}




\begin{proposition}[Probabilistic Modus Ponens (PMP)]
For all $\mathcal{EL}$ concepts $C, D, E$ (atomic or complex)
and all $\mathcal{SEL}$ interpretations $\interp$,
if $\tbox \models (D \mid C)[l_1, u_1]$ and
$\tbox \models (E \mid C \sqcap D)[l_2, u_2]$,
then $\tbox \models (E \mid C)[l_3, u_3]$,
where $l_3 = l_1 \cdot l_2$
and $u_3 = \min \{1, u_1 \cdot u_2 + 1 - l_1\}$.
\end{proposition}
\begin{proof}
Let us note that the proof works analogously for counting and volume semantics.
We give the proof for the latter, but by just replacing the volume with 
cardinality, we obtain the proof for counting semantics.

Consider an arbitrary $\mathcal{SEL}$ interpretation $\interp$ such 
that $\interp \models \tbox$. 
To prove the claim, we have to check that the lower and upper bounds in
\eqref{eq_sel_box_sat} hold for the conditional
$(E \mid C)[l_3, u_3]$.
For the lower bound, we have
\begin{align*}
\operatorname{Vol}(E^\interp \cap C^\interp) 
&=  \operatorname{Vol}(E^\interp \cap D^\interp \cap C^\interp) + \\
&\quad\operatorname{Vol}(E^\interp \cap (\neg D)^\interp\cap C^\interp) \\
&\geq l_2 \operatorname{Vol}(C^\interp \cap D^\interp) + 0\\
&\geq (l_2 \cdot l_1) \cdot \operatorname{Vol}(C^\interp),
\end{align*}
where we used $\tbox \models (E \mid C \sqcap D)[l_2, u_2]$ for the first
and $\tbox \models (D \mid C)[l_1, u_1]$ for the second inequality. 

Similarly, for the upper bound, we have 
\begin{align*}
\operatorname{Vol}(E^\interp \cap C^\interp) 
&=  \operatorname{Vol}(E^\interp \cap D^\interp \cap C^\interp) + \\
&\quad\operatorname{Vol}(E^\interp \cap (\neg D)^\interp\cap C^\interp) \\
&\leq u_2 \cdot \operatorname{Vol}(C^\interp \cap D^\interp)  + \\
&\quad\big(\operatorname{Vol}(C^\interp) - \operatorname{Vol}(C^\interp \cap D^\interp)   \big) \\
&\leq u_2 \cdot u_1 \cdot \operatorname{Vol}(C^\interp)  + \\
&\quad(\operatorname{Vol}(C^\interp) - l_1 \cdot \operatorname{Vol}(C^\interp)) \\
&= (u_2 \cdot u_1 + 1 - l_1) \cdot \operatorname{Vol}(C^\interp).
\end{align*}
Since
$\operatorname{Vol}(E^\interp \cap C^\interp ) \leq 1 \cdot \operatorname{Vol}(C^\interp)$, we
can also assume that $u_3 \leq 1$.
\end{proof}

\section{Encoding of Classical Axioms (BoxEL)}\label{app:boxel}

The encoding of the
four axiom types occurring in 
$\mathcal{EL}$ normal form in
\cite{XiongPTNS22} is based on
the \emph{disjoint measurement} defined by
$$
\contains(B_1, B_2) = 1-\frac{\operatorname{Vol}(B_1 \cap B_2)}{\operatorname{Vol}(B_1)}.
$$
The measure is guaranteed to be between
$0$ and $1$, is $0$ whenever $B_1 \subseteq B_2$ and is $1$ whenever $B_1 \cap B_2 = \emptyset$ \citep{XiongPTNS22},

\noindent
\textbf{NF1: Atomic Subsumption.} Axioms of the form $C \sqsubseteq D$ with
$D\neq \bot$ are encoded by the loss term
\begin{equation}\label{eq:loss_nf1}
    \mathcal{L}_{C \sqsubseteq D}(w) = \contains(\Box_w(C), \Box_w(D)).
\end{equation}
If $D=\bot$ and $C$ is not a nominal, that is, $C \sqsubseteq \bot$, the loss term
is
\begin{equation}\label{eq:loss_nf1}
    \mathcal{L}_{C \sqsubseteq \bot}(w) = 
    \max(0, M_w(C)_0 - m_w(C)_0 + \epsilon).
\end{equation}
If $C$ is a nominal, the axiom is inconsistent and this can be reported 
to the user.

\noindent
\textbf{NF2: Conjunctive Subsumption.} Axioms of the form 
$C \sqcap D \sqsubseteq E$ with $E \neq \bot$ are encoded by
the loss term 
\begin{equation}\label{eq:loss_nf2}
    \small
    \mathcal{L}_{C \sqcap D \sqsubseteq E}(w) = 
    \contains(\Box_w(C) \cap \Box_w(D), \Box_w(E)).
\end{equation}
For $E = \bot$, the loss term is
\begin{equation}\label{eq:loss_disjoint}
    \small
    \mathcal{L}_{C \sqcap D \sqsubseteq \bot}(w) = \frac{Vol(\Box_w(C)\cap \Box_w(D))}{Vol(\Box_w(C)) + Vol(\Box_w(D))}.
\end{equation}

\noindent
\textbf{NF3: Right Existential.}
Axioms of the form $C \sqsubseteq \exists r.D$. are encoded by
the loss term
\begin{equation}\label{eq:loss_nf3}
    \small\mathcal{L}_{C \sqsubseteq \exists r.D}(w) =
    \contains(T^r_w(\Box_w(C)), \Box_w(D)).
\end{equation}

\par

\noindent
\textbf{NF4: Left Existential.}
Axioms of the form $\exists r.C \sqsubseteq D$ are encoded by the loss term
\begin{equation}\label{eq:loss_nf4}
    \small\mathcal{L}_{\exists r.C \sqsubseteq D }(w) =
    \contains(T^{-r}_w(\Box_w(C)), \Box_w(D)),
\end{equation} 

As shown in \cite{XiongPTNS22},
it holds for all encodings that
$\mathcal{L}_\alpha(w) = 0$
implies that the corresponding
axiom $\alpha$ is satisfied by 
the geometric interpretation
$\interp_w$ corresponding to
the embedding.

\section{Further Experimental details}\label{app:impl}

\subsection{Softplus Volume and Cooling Schedule}\label{app:implementation}

We minimize the loss function by 
gradient descent. One practical problem 
is that if the intersection
of boxes is zero with respect to our
current parameter vector (embedding) $w$, it will still be $0$
in a small environment of $w$\footnote{If the distance between two boxes is $d$,
the corresponding components in $w$ have
to be changed by at least $d$ to let the
boxes intersect. For changes "smaller than d", they will still be disjoint.
}. This means that the gradient
will be $0$, which can prevent gradient
descent from bringing disjoint boxes closer 
together when necessary. This issue can be addressed
by approximating the volume with the softplus volume \citep{patel2020representing}
\begin{equation}\small\label{eq:softplus_volume}
 \operatorname{SVol}\left( \Box_w\left(C\right)\right)= \prod_{i=1}^d \operatorname{Softplus}_{t} \left( M_w\left(C  \right)_i - m_w\left(C\right)_i\right),
\end{equation}
where the \emph{softplus function} $\operatorname{Softplus}_{t}$ is defined as $\operatorname{softplus}_{t}(x)=t \log \left(1+e^{x/t}\right)$ and $t$ is called the \emph{temperature parameter}. Because of the exponential growth of the exponential function, we have $1+e^{x/t} \approx e^{x/t}$ when $x/t$ is large, so that $\operatorname{softplus}_{t}(x) \approx t \log \left(e^{x/t}\right) = x$. By letting $t$ go to $0$ as the search progresses (to let $x/t$ become bigger), the softmax volume will converge to the real volume as the algorithm progresses.
Our cooling schedule starts with a high temperature 
to learn fast at the beginning and decreases
the temperature gradually to converge to a proper minimum. 
We guarantee $M_w(C) \geq m_w(C)$ for all concepts $C \in \concepts$ by changing
the parametrization. Instead of 
using $M_w(C)$, we use a vector $\delta_w(C)$
of the same dimension and let $M_w(C) =
m_w(C) + \exp(\delta_w(C))$ (where the exponential function is applied element-wise).

\subsection{Details for SEL ontology construction}\label{app:data generation}

\textbf{Why Choose YAGO as the Knowledge Base?} 
YAGO3 is a knowledge base that combines the clean taxonomy of WordNet with the richness of the Wikipedia category system and assigns entities to more than 350,000 concepts. The accuracy of YAGO3 has been manually evaluated, with a confirmed accuracy of 95\%. We chose YAGO3 rather than DBPedia \citep{DBPdia} or Wikidata \citep{vrandevcic2014wikidata}, due to their issues in differentiating concepts and entities, as well as their rather noisy concept assertions. Furthermore, YAGO3 provides complete concept assertions for each entity, such as $Person(Claude Monet)$, which is automatically added to the knowledge base if $Artist\sqsubseteq Person$. This allows us to avoid counting incorrect numbers of entities due to missing concept assertions and create more accurate probabilities in the dataset.

\textbf{Concept \& Role Name Selection:}
We created the conditionals by choosing subsets of concepts and roles from a particular domain. 
For the \emph{Country} and \emph{Person} dataset, we concentrate on concepts and roles belonging to $\langle\emph{wordnet\_country\_108544813}\rangle$
and $\langle\emph{wordnet\_person\_100007846}\rangle$, respectively. The \emph{Hybrid} dataset is crafted by selecting a set of roles that bridge different domains and choosing the relevant top-level concepts from each domain. Specifically, roles are ranked by their fact count, with the top 18 selected. Concepts in the domain or range of these roles are then ranked by instance count, and the top 12 concepts are chosen.

\textbf{Conditional Construction:}
We systematically traverse all conditionals of type PNF1 - PNF4 that can be generated from the selected concepts and roles and define their probabilities by statistical proportions. Algorithm \ref{code:data} provides a detailed explanation of the process used to obtain probabilities for the conditionals.

\begin{algorithm}[t]
\caption{Pseudocode for the data generation}
\label{code:data}
\begin{algorithmic}[1]
    \STATE $\mathcal{L}_{concept}\leftarrow$ select concepts in certain domain.
    \STATE $\mathcal{L}_{role}\leftarrow$ select roles correspondingly.
    \vspace{8pt}
    \STATE /* Generate PNF1 */
    \STATE $\mathcal{D}_{PNF1}\leftarrow$ Empty List
    \FORALL{$A\in\mathcal{L}_{concept}$}
        \FORALL{$B\in\mathcal{L}_{concept}/A$}
            \STATE $n_{A}\leftarrow$ count the number of entities belonging to $A$
            \STATE $n_{A\sqcap B}\leftarrow$ count the number of entities belonging to both $A$ and $B$
            \STATE $p\leftarrow\frac{n_{A\sqcap B}}{n_{A}}$
            \STATE Append $(B|A)[p]$ to $\mathcal{D}_{PNF1}$
        \ENDFOR
    \ENDFOR
    \vspace{8pt}
    \STATE /* Generate PNF2 */
    \STATE $\mathcal{D}_{PNF2}\leftarrow$ Empty List
    \FORALL{$A_1\in\mathcal{L}_{concept}$}
        \FORALL{$A_2\in\mathcal{L}_{concept}/A_1$}
            \FORALL{$B\in\mathcal{L}_{concept}/\{A_1,A_2\}$}
                \STATE $n_{A_1\sqcap A_2}\leftarrow$ count the number of entities belonging to both $A_1$ and $A_2$
                \STATE $n_{A_1\sqcap A_2\sqcap B}\leftarrow$ count the number of entities belonging to all $A_1$, $A_2$ and $B$
                \STATE $p\leftarrow\frac{n_{A_1\sqcap A_2\sqcap B}}{n_{A_1\sqcap A_2}}$
                \STATE Append $(B|A_1\sqcap A_2)[p]$ to $\mathcal{D}_{PNF2}$
            \ENDFOR
        \ENDFOR
    \ENDFOR
    \vspace{8pt}
    \STATE /* Generate PNF3 and PNF4*/
    \STATE $\mathcal{D}_{PNF3}\leftarrow$ Empty List
    \STATE $\mathcal{D}_{PNF4}\leftarrow$ Empty List
    \FORALL{$r\in\mathcal{L}_{role}$}
        \FORALL{$A\in\mathcal{L}_{concept}$}
            \FORALL{$B\in\mathcal{L}_{concept}/A$}
                \STATE $n_{B}\leftarrow$ count the number of entities belonging to $B$
                \STATE $n_{\exists r.A}\leftarrow$ count the number of entities related to other entities belonging to $A$ through the role name $r$
                \STATE $n_{B\sqcap \exists r.A}\leftarrow$ count the number of entities belonging to $B$ related to other entities belonging to $A$ through the role name $r$
                \STATE $p_{PNF3}\leftarrow\frac{n_{B\sqcap \exists r.A}}{n_{\exists r.A}}$
                \STATE Append $(B|\exists r.A)[p_{PNF3}]$ to $\mathcal{D}_{PNF3}$
                \STATE $p_{PNF4}\leftarrow\frac{n_{B\sqcap \exists r.A}}{n_{B}}$
                \STATE Append $(\exists r.A|B)[p_{PNF4}]$ to $\mathcal{D}_{PNF4}$
            \ENDFOR
        \ENDFOR
    \ENDFOR
\end{algorithmic}
\end{algorithm}

\subsection{Pseudocode for the PMP evaluation}\label{app:PMP}
We denote $(Q_2|Q_1)[p]$ as a probabilistic conditional in our entire axiom set $\mathcal{D}$, where $Q_1$ (body) and $Q_2$ (head) could be arbitrary $\mathcal{EL}$ concepts. We further define $\mathcal{D}'\subset\mathcal{D}$ as a set of conditionals whose body is the most general concept in knowledge base, i.e., $\langle wordnet\_person \rangle$ for "Person" knowledge base. We randomly select 30\% of conditionals from $\mathcal{D}'$ as our query set $\mathcal{D}_{query}$. $\mathcal{D}_{model}$ refers to the set of axioms that are used to learn embeddings.

Following algorithm ensures that
\begin{itemize}
    \item our knowledge base for learning embeddings does not contain $(Q_2|Q_1)[p]$.
    \item Our knowledge base for learning embeddings contains the premisses required for PMP evaluation, i.e., $(A|Q_1)[q1]$ and $(Q_2|A)[q2]$.
\end{itemize}

\begin{algorithm}[t]
\caption{Pseudocode for the PMP Evaluation}
\label{code:PMP}
\begin{algorithmic}[1]
\STATE /* prepare knowledge base for learning embeddings and the query set */ 
\STATE $\mathcal{D}_{query}\leftarrow RandomSample(\mathcal{D}', 30\%)$
\STATE $\mathcal{D}_{model}\leftarrow Empty\ List$
\FORALL{$(Q_2|Q_1)[p]\in\mathcal{D}$}
    \IF{$D\not\in\mathcal{D}_{query}$}
        \STATE Append $(Q_2|Q_1)[p]$ to $\mathcal{D}_{model}$
    \ENDIF
\ENDFOR
\vspace{8pt} 

\STATE /* Knowledge base modeling */ 
\STATE Learning the embeddings $\mathcal{M}$ with $\mathcal{D}_{model}$
\vspace{8pt} 

\STATE /* PMP evaluation */
\STATE $BD\leftarrow 0, Hits\leftarrow 0, n_{eval}\leftarrow 0$
\FORALL{$(Q_2|Q_1)[p]\in \mathcal{D}$}
    \STATE $n_{eval}\leftarrow n_{eval}+1$
    \IF{$(Q_2|Q_1)[p]\in \mathcal{D}_{query}\wedge (A|Q_1)[q1]\in\mathcal{D}_{model}\wedge(Q_2|A)[q2]\in\mathcal{D}_{model}$}
        \STATE /* PMP evaluation */
        \STATE $Bound_{min}\leftarrow q1*q2$
        \STATE $Bound_{max}\leftarrow \min(1, q1*q2+1-q2)$
        \vspace{3pt}
        \STATE $\overline{p}\leftarrow\mathcal{M}((Q_2|Q_1))$
        \vspace{3pt}
        \IF{$\overline{p}<Bound_{min}\lor \overline{p}>Bound_{max}$}
        \vspace{3pt}
        \STATE $BD\leftarrow BD + \max(0, Bound_{min}-\overline{p})+\max(0, \overline{p}-Bound_{max})$
        \vspace{3pt}
        \ELSE
        \STATE $Hits\leftarrow Hits +1$
        \ENDIF
    \ENDIF
\ENDFOR
\vspace{8pt}
\STATE $SE\leftarrow BD/n_{eval}$
\STATE $SA\leftarrow Hits/n_{eval}*100\%$
\RETURN $SE$, $SA$
\end{algorithmic}
\end{algorithm}

\begin{table}[h!]
\centering
\resizebox{.45\textwidth}{!}{
\begin{tabular}{ccccc}
\toprule
                                            &      & \textbf{\# Total}  & \textbf{\# Knowledge Base} & \textbf{\# Query}  \\ \midrule
\multicolumn{1}{c}{\multirow{4}{*}{\textbf{Person}}} & PNF1 & 16002  & 10962 & 3480  \\
\multicolumn{1}{c}{}                        & PNF2 & 117600 & 75264 & 17856 \\
\multicolumn{1}{c}{}                        & PNF3 & 96012  & 65772 & 9054  \\
\multicolumn{1}{c}{}                        & PNF4 & 96012  & 67284 & 8316  \\\midrule
\multirow{4}{*}{\textbf{Country}}                    & PNF1 & 870    & 638   & 176   \\
                                            & PNF2 & 24360  & 17864 & 3696  \\
                                            & PNF3 & 870    & 638   & 70    \\
                                            & PNF4 & 870    & 609   & 58    \\\midrule
\multirow{4}{*}{\textbf{Hybrid}}                     & PNF1 & 156    & 120   & 30    \\
                                            & PNF2 & 1716   & 1320  & 270   \\
                                            & PNF3 & 2808   & 2160  & 194   \\
                                            & PNF4 & 2808   & 1968  & 192   \\ \bottomrule
\end{tabular}
}
\caption{Statistics of the knowledge base/query set split for the PMP evaluation}\label{tab:train_test}
\end{table}

\subsection{Details for KDE Estimator}\label{app:kde_estimator}
Let $\mathbf{I_1}, \mathbf{I_2}, \dots, \mathbf{I_m}$ denote the probability intervals of the conditionals in the learning set, where each interval $\mathbf{I_i} = [l_i, u_i]$. These intervals can be treated as samples from a 2-variate random variable with a probability density function $f$. To estimate $f$, we use kernel density estimation, defined as
\begin{equation}
    \bar f_\mathbf{H}(\mathbf{I}) = \frac{1}{m}\sum_{i=1}^m K_{\mathbf{H}}(\mathbf{I}-\mathbf{I_i}),
\end{equation}
where $K_{\mathbf{H}}$ is the kernel function.

In our baseline, we employ Gaussian kernel to estimate the density:
\begin{equation}
    K_{\mathbf{H}}(\mathbf{I}-\mathbf{I_i})=\frac{1}{\mathbf{|H|}^{1/2}\sqrt{2\pi}}\exp{(-\frac{1}{2}(\mathbf{I}-\mathbf{I_i})^{\top}\mathbf{H}^{-1}(\mathbf{I}-\mathbf{I_i}))},
\end{equation}

where $\mathbf{H}$ is the bandwidth, a symmetric and positive definite $m\times m$ matrix that balances smoothness and details in the density estimate. To select the bandwidth, we apply Scott's rule \citep{scott2015multivariate}. For a 2-variate density estimate, the bandwidth matrix is given by:
\begin{equation}
    \mathbf{H}=diag(\sigma_1, \sigma_2)\cdot m^{-\frac{1}{6}},
\end{equation}
where $\sigma_1$, $\sigma_2$ are the standard deviation of $l_i$ and $u_i$, respectively.

We then generate predicted intervals $[\bar l,\bar u]$ for KDE estimator by sampling from the estimated density function $\bar f_\mathbf{H}(\mathbf{I})$.

\section{Additional Experimental Results}\label{app:moreresults}
\subsection{Hyperparameter Tuning}\label{app:hyper_param}
We tune the main hyperparameters separately for each ontology and report the selected values in Table~\ref{tab:hyper_param}. To assess sensitivity, we vary one hyperparameter at a time around the selected configuration and report the mean and standard deviation of the resulting performance in Table~\ref{tab:hyperparameter_sensitivity}. Overall, the results indicate that performance is relatively stable across the tested ranges.

\begin{table}[h!]
    \centering
    \resizebox{0.45\textwidth}{!}{%
    \begin{tabular}{cccccc}
    \toprule
    & ED & EP & BS & LR & BR \\\midrule
    Person & 32 & 20 & 512 & 0.01 & [0,10]\\
    Hybrid & 8 & 20 & 256 & 0.01 & [0,1]\\
    Country & 16 & 30 & 256 & 0.05 & [0,10]\\
    \bottomrule
    \end{tabular}
    }
    \caption{Best hyperparameters used for different ontologies: embedding dimension (ED), number of epochs (EP), batch size (BS), learning rate (LR),  bound for regularization (BR).}
    \label{tab:hyper_param}
\end{table}

\begin{table}[h!]
\centering
\begin{tabular}{lcccc}
\toprule
\textbf{Hyperparameter} & \textbf{Search Range} & \textbf{Person} & \textbf{Country} & \textbf{Hybrid} \\
\midrule
Embedding dim $n$ & $\{8, 16, 32, 64, 128\}$ & $0.088 \pm 0.052$ & $0.042 \pm 0.032$ & $0.065 \pm 0.025$ \\
Learning rate $\mathrm{lr}$ & $\{0.001, 0.01, 0.05, 0.1\}$ & $0.080 \pm 0.025$ & $0.048 \pm 0.021$ & $0.075 \pm 0.027$ \\
Side length $\beta$ & $\{1, 10\}$ & $0.120 \pm 0.070$ & $0.030 \pm 0.010$ & $0.065 \pm 0.025$ \\
\bottomrule
\end{tabular}
\caption{Hyperparameter search ranges and corresponding results across datasets.}
\label{tab:hyperparameter_sensitivity}
\end{table}

\subsection{Fine-grained Approximate Gap Evaluation}
\begin{table}[h!]
\centering
\resizebox{.45\textwidth}{!}{
\begin{tabular}{llllll}
\toprule
\multicolumn{1}{c}{\multirow{2}{*}{Datasets}} & \multicolumn{5}{c}{Approximation Gap} \\
\cmidrule(lr){2-6} 
\multicolumn{1}{c}{} &
  \multicolumn{1}{c}{Total} &
  \multicolumn{1}{c}{PNF1} &
  \multicolumn{1}{c}{PNF2} &
  \multicolumn{1}{c}{PNF3} &
  \multicolumn{1}{c}{PNF4} \\ \midrule
PERSON & 0.23 & 0.22 & 0.21 & 0.30 & 0.22 \\
COUNTRY & 0.25 & 0.24 & 0.24 & 0.28 & 0.24 \\
HYBRID & 0.29 & 0.29 & 0.27 & 0.33 & 0.27  \\ \bottomrule
\end{tabular}
}
\caption{Approximation gap evaluated on 60 embeddings.}
\label{tab:interval_result}
\end{table}

\newpage
\section{Ablation Studies}\label{app:ablation}
\begin{table}[h!]
\centering
\resizebox{0.5\textwidth}{!}{%
{\small
\begin{tabular}{ccccc}
\toprule[1.2pt]
Dataset & Concept Representation & SA $\uparrow$ & SE $\downarrow$ & AG $\downarrow$ \\\midrule
 & n-balls & 85.7\% & 0.022 & 0.324 \\
  \multirow{-2}{*}{Person} & boxes$^*$ & \textbf{89.3\%} & \textbf{0.013} & \textbf{0.233} \\\midrule
 & n-balls & 90.3\% & 0.018 & 0.359 \\
  \multirow{-2}{*}{Country} & boxes$^*$ & \textbf{95.6\%} & \textbf{0.004} & \textbf{0.241} \\\midrule
 & n-balls & 82.9\% & 0.052 & 0.344 \\
  \multirow{-2}{*}{Hybrid} & boxes$^*$ & \textbf{89.7\%} & \textbf{0.020} & \textbf{0.293} \\\bottomrule[1.2pt]
\end{tabular}%
}
}
\caption{Ablation study of the concept representation. The design choice of our method is marked with an asterisk (*).}
\label{tab:abl_concept}
\end{table}

\begin{table}[h!]
\centering
\resizebox{0.5\textwidth}{!}{%
{\tiny
\begin{tabular}{ccccc}
\toprule[0.7pt]
Dataset & Regularization Term & SA $\uparrow$ & SE $\downarrow$ & AG $\downarrow$ \\\midrule
 & w/o & 45.1\% & 0.220 & \textbf{0.202} \\
 & $\mathcal{L}_{loc}$ & 85.7\% & 0.150 & 0.276 \\
 & $\mathcal{L}_{vol}$ & 87.3\% & 0.015 & 0.235 \\
\multirow{-4}{*}{Person} & $(\mathcal{L}_{loc}+\mathcal{L}_{vol})^*$ & \textbf{89.3\%} & \textbf{0.013} & 0.233 \\\midrule
 & w/o & 82.1\% & 0.090 & 0.243 \\
 & $\mathcal{L}_{loc}$ & 90.2\% & 0.110 & 0.255 \\
 & $\mathcal{L}_{vol}$ & 95.1\% & 0.080 & 0.253 \\
\multirow{-4}{*}{Country} & $(\mathcal{L}_{loc}+\mathcal{L}_{vol})^*$ & \textbf{95.60\%} & \textbf{0.004} & \textbf{0.241} \\\midrule
 & w/o & 86.2\% & 0.035 & 0.297 \\
 & $\mathcal{L}_{loc}$ & 87.3\% & 0.018 & 0.319 \\
 & $\mathcal{L}_{vol}$ & 88.2\% & \textbf{0.002} & 0.301 \\
\multirow{-4}{*}{Hybrid} & $(\mathcal{L}_{loc}+\mathcal{L}_{vol})^*$ & \textbf{89.70\%} & \textbf{0.002} & \textbf{0.293}\\\bottomrule[0.7pt]
\end{tabular}%
}
}
\caption{Ablation study of the regularization terms. The design choice of our method is marked with an asterisk (*).}
\label{tab:abl_reg}
\end{table}

\end{document}